%% file: main.tex
\documentclass[a4paper,conference]{IEEEtran}
\IEEEoverridecommandlockouts

\ifCLASSINFOpdf
\else
\fi
\usepackage{graphicx} 
\usepackage{cite}
\usepackage{epsfig}
\usepackage{subcaption}
\usepackage{caption}
\usepackage{algorithm}
\usepackage{algorithmic}
\usepackage{newfloat}
\usepackage{listings}
\usepackage{booktabs}
\usepackage{multicol}
\usepackage{tabularx}
\usepackage{amsmath}
\usepackage{array}
\usepackage{bibentry}
\usepackage{amsmath}
\usepackage[dvipsnames]{xcolor}
\usepackage{multirow}
\usepackage{amsfonts}
\usepackage{pgfplots}
\DeclareUnicodeCharacter{2212}{−}
\usepgfplotslibrary{groupplots,dateplot}
\usetikzlibrary{patterns,shapes.arrows}
\pgfplotsset{compat=newest}

\begin{document}
%
\title{Learning Fashion Compatibility \\ from In-the-wild Images}

\author{\IEEEauthorblockN{Additya Popli}
\IEEEauthorblockA{International Institute of Information and Technology,\\
Hyderabad, India 500032\\
Email: additya.popli@research.iiit.ac.in}
\and
\IEEEauthorblockN{Vijay Kumar}
\IEEEauthorblockA{Walmart Global Tech, India\\
Email: vij4321r@gmail.com}
\and
\IEEEauthorblockN{Sujit Jos}
\IEEEauthorblockA{Walmart Global Tech, India\\
Email: sujitjos1729@gmail.com}
\and
\IEEEauthorblockN{Saraansh Tandon}
\IEEEauthorblockA{International Institute of Information and Technology,\\
Hyderabad, India 500032\\
Email: saraansh.tandon@research.iiit.ac.in}}


%
\author{\IEEEauthorblockN{Additya Popli\IEEEauthorrefmark{1},
Vijay Kumar\IEEEauthorrefmark{2},
Sujit Jos\IEEEauthorrefmark{2} and
Saraansh Tandon\IEEEauthorrefmark{1}}
\IEEEauthorblockA{\IEEEauthorrefmark{1}International Institute of Information Technology, Hyderabad, India \\ Email: \{additya.popli, saraansh.tandon\}@research.iiit.ac.in}
\IEEEauthorblockA{\IEEEauthorrefmark{2}Email: \{vij4321r, sujitjos1729\}@gmail.com}}


\maketitle

\begin{abstract}
\input{section/abstract}
\end{abstract}


%
\IEEEpeerreviewmaketitle

\section{Introduction}
\label{sec:intro}

\input{section/introduction}

\section{Related Work}
\label{sec:related_work}
\input{section/related_work}

\section{Proposed Approach}
\label{sec:main}
\input{section/main}

\section{Datasets}
\label{sec:datasets}
\input{section/datasets}

\section{Experiments}
\label{sec:experiments}
\input{section/experiments}

\section{Conclusion}
\input{section/conclusion}

\bibliographystyle{IEEEtran}
\bibliography{IEEEabrv, IEEEexample}

\end{document}


%
\title{Learning Fashion Compatibility \\ from In-the-wild Images \\ Supplementary Material}

\author{\IEEEauthorblockN{Additya Popli}
\IEEEauthorblockA{International Institute of Information and Technology,\\
Hyderabad, India 500032\\
Email: additya.popli@research.iiit.ac.in}
\and
\IEEEauthorblockN{Vijay Kumar}
\IEEEauthorblockA{Walmart Global Tech, India\\
Email: vij4321r@gmail.com}
\and
\IEEEauthorblockN{Sujit Jos}
\IEEEauthorblockA{Walmart Global Tech, India\\
Email: sujitjos1729@gmail.com}
\and
\IEEEauthorblockN{Saraansh Tandon}
\IEEEauthorblockA{International Institute of Information and Technology,\\
Hyderabad, India 500032\\
Email: saraansh.tandon@research.iiit.ac.in}}


%
\author{\IEEEauthorblockN{Additya Popli\IEEEauthorrefmark{1},
Vijay Kumar\IEEEauthorrefmark{2},
Sujit Jos\IEEEauthorrefmark{2} and
Saraansh Tandon\IEEEauthorrefmark{1}}
\IEEEauthorblockA{\IEEEauthorrefmark{1}International Institute of Information Technology, Hyderabad, India \\ Email: \{additya.popli, saraansh.tandon\}@research.iiit.ac.in}
\IEEEauthorblockA{\IEEEauthorrefmark{2}Email: \{vij4321r, sujitjos1729\}@gmail.com}}

\maketitle

\IEEEpeerreviewmaketitle

\section{Qualitative Analysis}

\begin{figure}[t]
\centering
\fbox{
\begin{subfigure}[]{}
\centering
\includegraphics[width=\columnwidth]{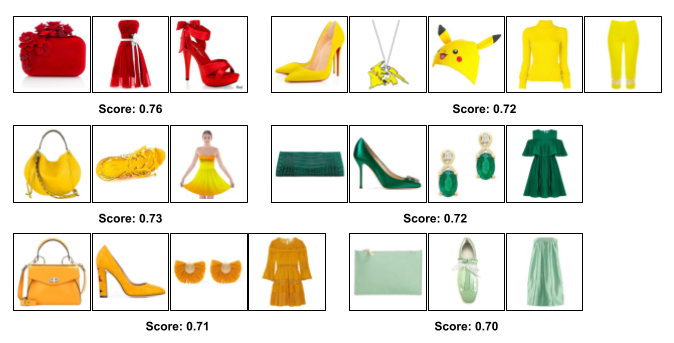}
\label{fig:supp_compatible_outfits}
\end{subfigure}}%

\fbox{
\begin{subfigure}[]{}
\centering
\includegraphics[width=\columnwidth]{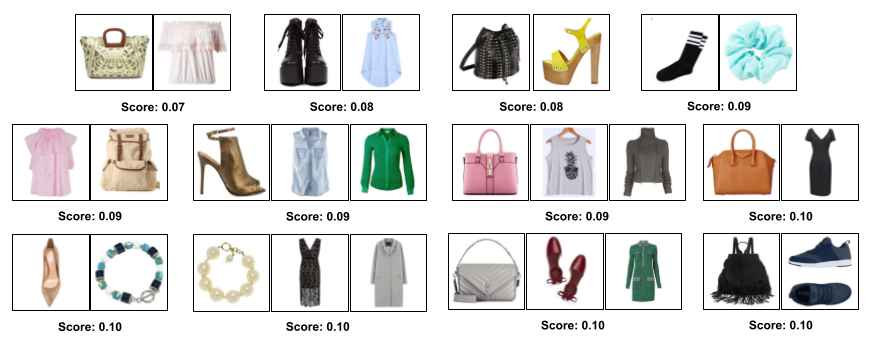}
\label{fig:supp_noncompatible_outfits}
\end{subfigure}}
\caption{Few examples showing compatible (a) and non-compatible (b) outfits correctly predicted by our model}
\label{fig:supp_compatible_noncompatible}
\end{figure}

\begin{figure}
\centering
\begin{subfigure}{}
\fbox{\includegraphics[width=\columnwidth]{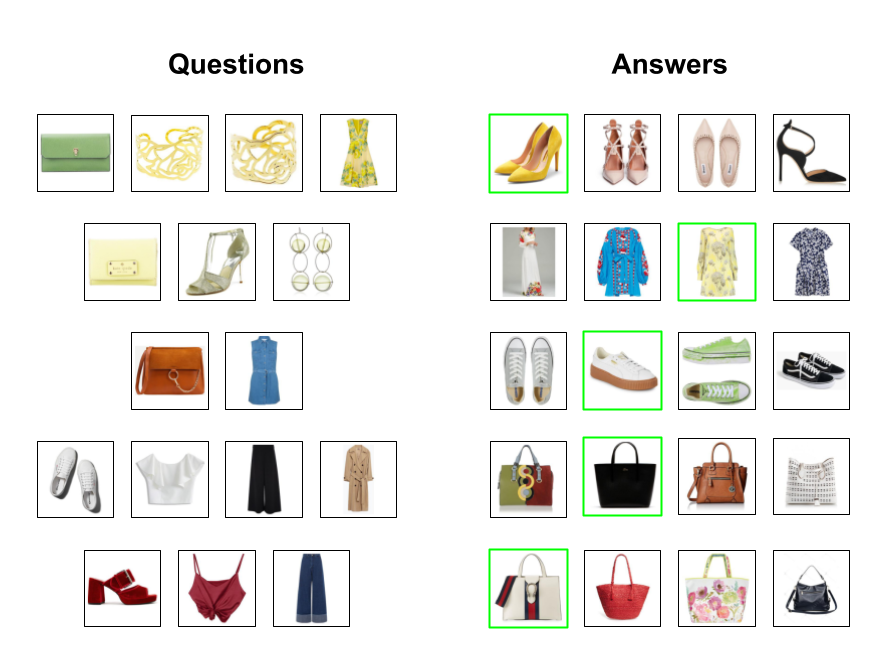}}
\label{fig:supp_fitb_success}
\end{subfigure}
\begin{subfigure}{}
\fbox{\includegraphics[width=\columnwidth]{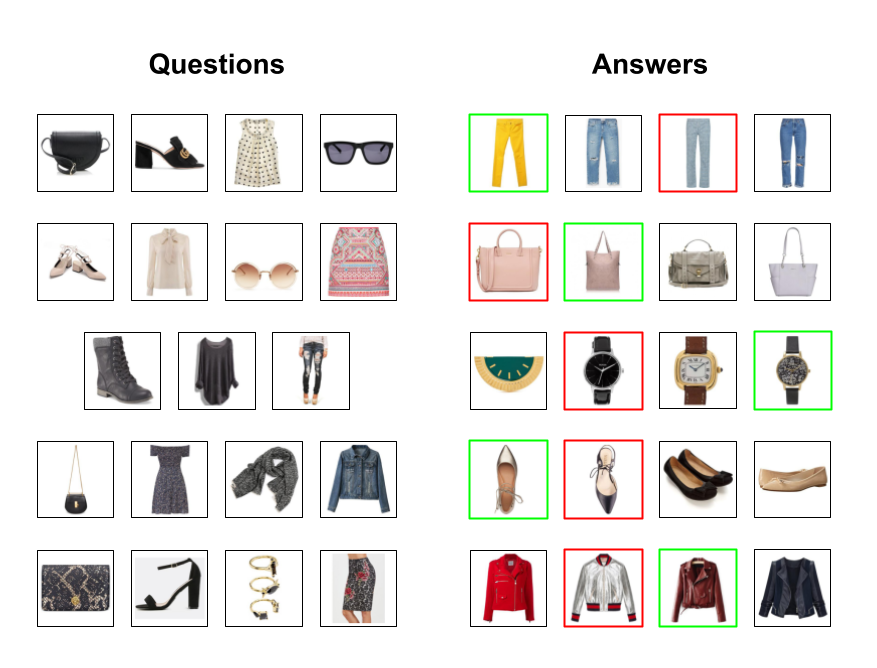}}
\label{fig:supp_fitb_wrongs}
\end{subfigure}
\caption{Success (top) and failure (bottom) examples for Fill-in-the-blank task on Polyvore dataset. (top) Few examples where our model successfully predicted the right candidate where green box indicate the correct candidate selected by our model. (bottom) Few examples where our model predicted the wrong candidate where green and red boxes indicate ground-truth and predicted candidates, respectively.}
\label{fig:supp_fitb}
\end{figure}

We show a few success and failure cases of our model on the compatibility and fill-in-the-blank tasks in Figure~\ref{fig:supp_compatible_noncompatible} and  Figure~\ref{fig:supp_fitb}, respectively. Notice the failure cases in Figure~\ref{fig:supp_fitb} that contain annotation ambiguities (e.g. 2nd and 3rd rows in bottom figure) where more than one candidate seem to be the right choice. It clearly shows the limitation of the existing evaluation protocol that needs be addressed.

\begin{figure}
\centering
\includegraphics[width=\columnwidth]{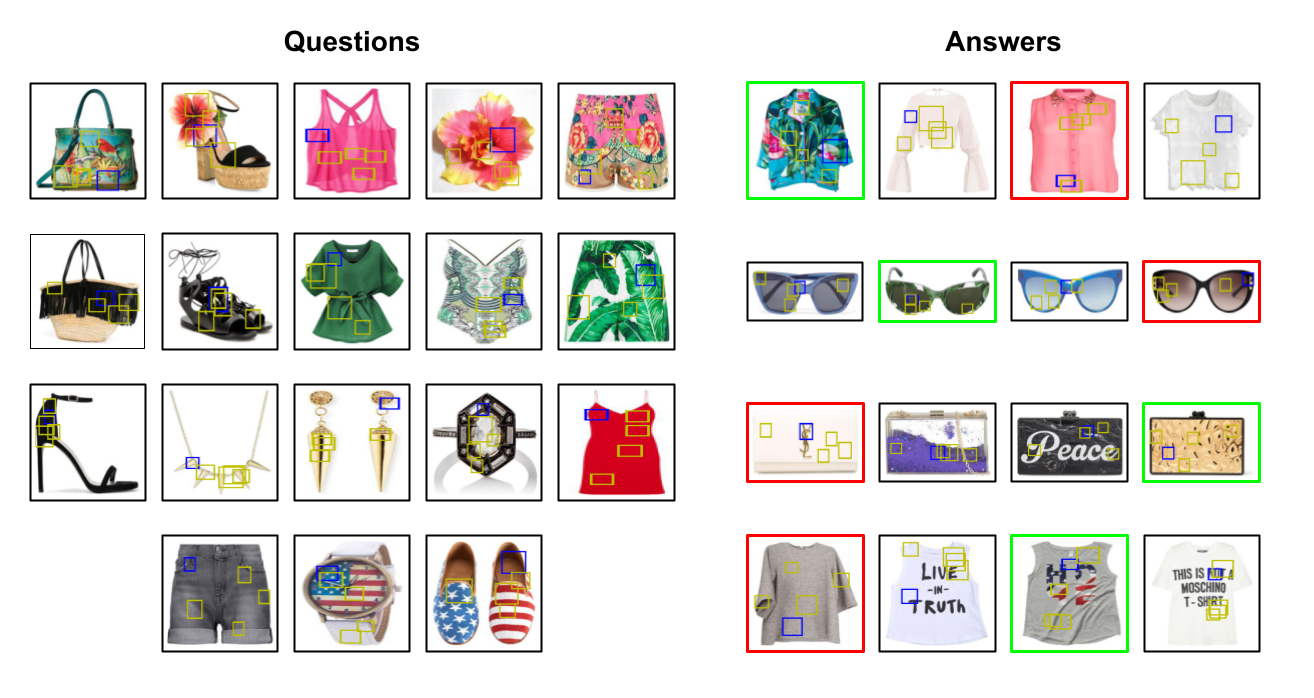}
\caption{Effectiveness of multiple patches (yellow) during inference compared to single patch (blue). Green and red boxes around the candidates indicate the correct/incorrect prediction made by our model.}
\label{fig:supp_role_of_patches}
\end{figure}

In Figure~\ref{fig:supp_role_of_patches} we show the benefit of using multiple patches during inference. Predictions made using multiple patches (yellow) are more robust compared to those made using only a single patch (blue) as random selection might capture unwanted regions around edges, corners or other local patterns that are not representative of the item.

\begin{figure*}
\centering
\includegraphics[width=\linewidth]{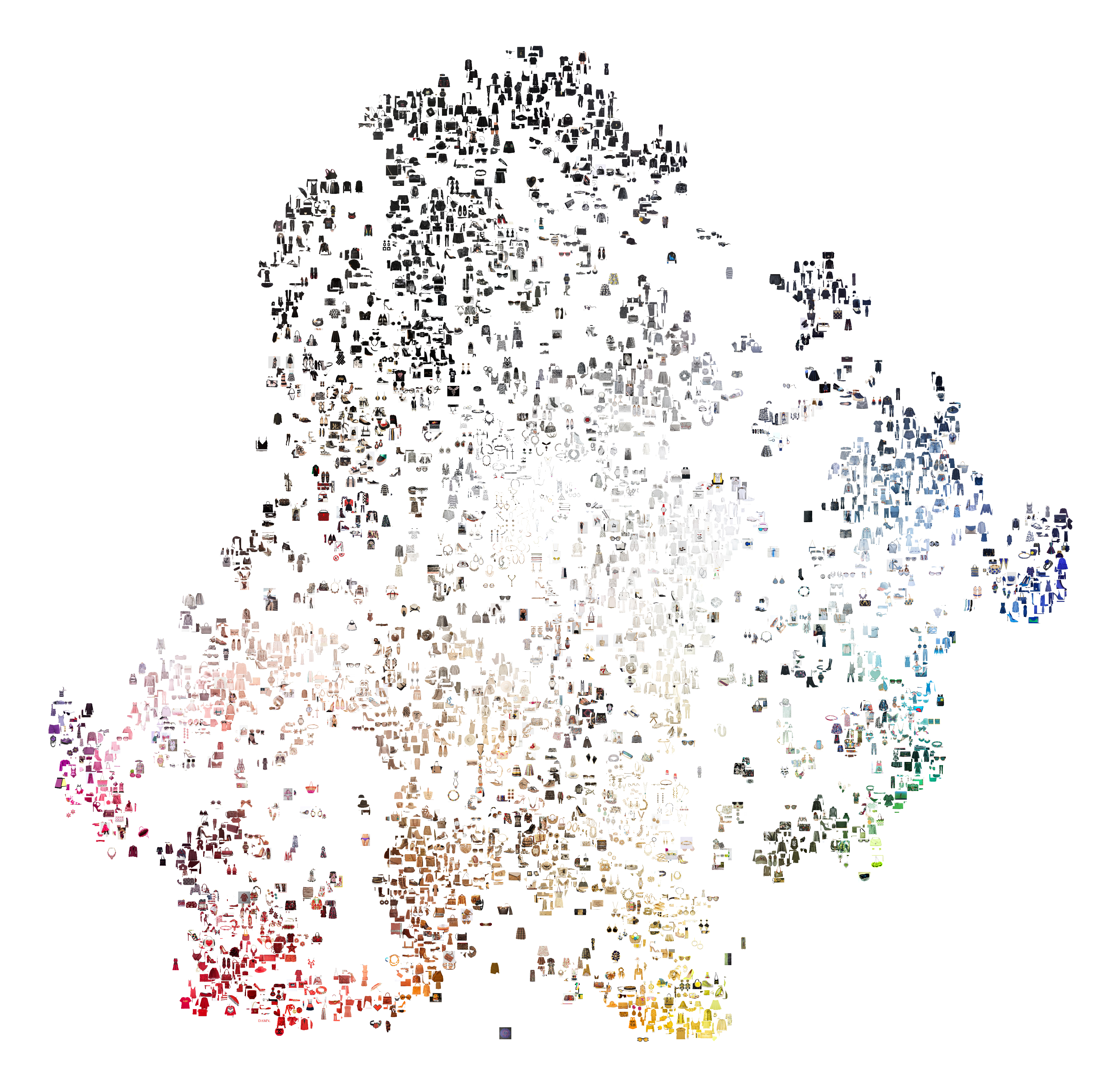}
\caption{t-sne visualization of Polyvore dataset on embeddings obtained by our model.}
\label{fig:supp_tsne_supplementary}
\end{figure*}

Finally, in Figure~\ref{fig:supp_tsne_supplementary} we show the t-sne visualization of the Polvore dataset on embeddings obtained by our proposed model. Unlike \cite{kim2020self} that learns colour and texture similarities, our embeddings trained for fashion compatibility result in compatible items that belong to different categories to cluster together tightly. Interestingly, our model is also able to capture colour and textures attributes without explicit loss functions introduced in \cite{kim2020self} to learn those attributes.

\bibliographystyle{IEEEtran}
\bibliography{IEEEabrv, IEEEexample}


%% file: section/abstract.tex
Complementary fashion recommendation aims at identifying items from different categories ({\em e.g.} shirt, footwear, {\em etc}.) that ``go well together'' as an outfit. Most existing approaches learn representation for this task using labeled outfit datasets containing manually curated compatible item combinations. In this work, we propose to learn representations for compatibility prediction from in-the-wild street fashion images through self-supervised learning by leveraging the fact that people often wear compatible outfits. Our pretext task is formulated such that the representations of different items worn by the same person are closer compared to those worn by other people. Additionally, to reduce the domain gap between in-the-wild and catalog images during inference, we introduce an adversarial loss that minimizes the difference in feature distribution between the two domains. We conduct our experiments on two popular fashion compatibility benchmarks - Polyvore and Polyvore-Disjoint outfits, and outperform existing self-supervised approaches, particularly significant in cross-dataset setting where training and testing images are from different sources.

%% file: section/introduction.tex
With the steady growth in online shopping, recommendation systems have become critical to drive customer engagement and revenue in many e-commerce applications. In this work, we consider complementary fashion recommendation where the objective is to determine whether a given set of items are compatible - predicting if the items from different categories ({\em e.g.} shirt, trouser, shoes, {\em etc.,}) can be worn together in an outfit. The problem is challenging as it not only requires understanding of colour, texture in individual items as well as outfits, but also higher level subjective reasoning about style, aesthetics and trend.

\begin{figure}[h]
\begin{center}
\frame{\includegraphics[width=0.75\columnwidth]{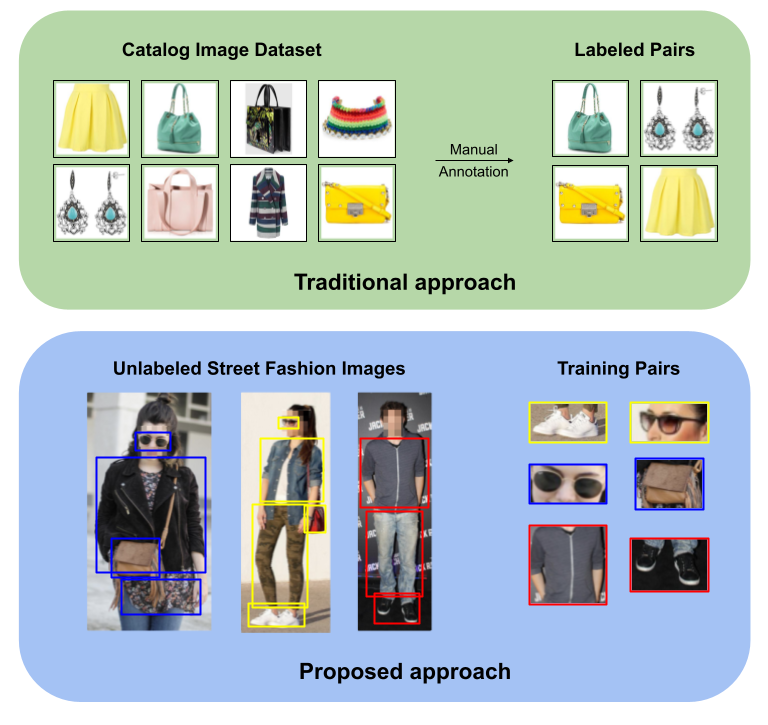}}
\end{center}
\caption{Existing approaches (top) train compatibility models on large labeled datasets that are curated through exhaustive annotation of catalog item combinations. On the contrary, the proposed approach (bottom) learns the embeddings from photos of people wearing different outfits in the wild images.} 
\label{fig:abstract}
\end{figure}

Learning representations for the compatibility prediction task is straight-forward if a labeled outfit dataset is available. Metric learning techniques \cite{eccv2018learning, iccv2019learning} can then be employed to learn an embedding space where compatible (positive) pairs from different categories are closer compared to non-compatible (negative) item pairs. However, creating such a labeled dataset is cumbersome, expensive, and often not feasible for a large product catalog to go exhaustively over all possible item combinations (see Figure~\ref{fig:abstract} (top)). Apart from scale, labeling is also prone to human biases as fashion preferences are subjective and models might learn these biases if not handled properly. Another option is to mine user co-purchase patterns, however that often leads to noisy outfits with high false positive rates \cite{hsiao2018creating}.

In this work, we focus on learning representations for complementary fashion recommendation by leveraging in-the-wild street fashion images uploaded by people. These images are easily available, scalable and reflect true choices and preferences of outfit combinations, thus are less biased. Learning directly from street images provides a flexibility to adopt the models to changing fashion trends and seasonality. 

Given a corpus of street fashion images as shown in Figure~\ref{fig:abstract} (bottom), we formulate the problem as a self-supervised task where the objective is to learn the embeddings such that the representations of different items worn by the same person are closer compared to those worn by a different person. Since shapeless features are required to measure the closeness between different categories, we resort to patch based representations unlike in previous approaches that focus on image based representations. Additionally, patch based representation avoid estimation of unknown transformations for different viewpoint, occlusion and warping of fashion items seen in real world images. As shown in our experiments on different datasets, patch representations are as effective as image representation for fashion compatibility.

There is a significant domain difference between user photos and catalog images that are our domain of interest. Unlike street images, catalog images are captured in a controlled setting with uniform background and usually with high illumination and resolution compared to street images. To address this limitation, we train our model in an adversarial learning framework iterating over representation learning from the street images, and enabling knowledge transfer to catalog images through domain adaption. 


We conduct extensive evaluation of our proposed approach on various compatibility prediction tasks on two popular benchmarks Polyvore and Polyvore-D \cite{eccv2018learning} datasets. Our approach performs better than state-of-the-art self-supervised approaches, and outperform significantly especially on cross-dataset evaluation in which training and test images are collected from different sources. To the best of our knowledge, ours is the first attempt to learn image representations from in-the-wild street images from fashion compatibility task.

%% file: section/related_work.tex

Fashion Compatibility is first attempted as a sequential problem using a bi-directional \textsc{LSTM} \cite{bilstm} to predict the next compatible item from given outfit items. Following work \cite{eccv2018learning} propose a metric learning approach to learn an embedding space for every pair of item categories. However, the method is not scalable to catalogs with large number of categories. In another work, attention masks are introduced to learn several fashion attributes such as colour and texture explicitly \cite{iccv2019learning}. A ranking based loss function that optimizes distances across all outfit item pairs rather than a pair of triplets is proposed in \cite{cvpr2020fashion}. More recently, graph-based solutions \cite{yang2020learning, cucurull2019context, Duannips} that model the relationship within a outfit graph have shown to outperform previous approaches. 


Self-Supervised Learning (SSL) has become an important sub-field in machine learning to address the scarcity of large labeled datasets \cite{hinton1993autoencoders, zhang2016colorful,gidaris2018unsupervised, noroozi2016unsupervised}. 
Kim {\it et al.,} \cite{kim2020self} aimed to learn representations for compatibility prediction by solving self-supervised tasks to learn colour, texture and shape features explicitly. Although their method outperforms several standard self-supervised solutions, the performance drops significantly in a cross-dataset setting as they rely on similarity rather compatibility between pairs of image patches. The performance drops when the model is trained on one dataset and test on different dataset of catalog images.

Contrary to \cite{kim2020self} that learn visual similarity, we focus on learning patch compatibility from in-the-wild street fashion photos with self-supervision. The proposed method is similar to \cite{hsiao2018creating} that also leverage in-the-wild images. However, \cite{hsiao2018creating} observed a large domain gap between catalog and street fashion images, and hence focused on clothing attribute labels rather than visual features. In our work, we directly address the domain gap between source and target domains with adversarial training. 

Adversarial learning is another area closely related to ours. It is widely applied in many applications from generating realistic images \cite{goodfellow2014generative, simGAN}, domain adaption \cite{GAN_domain_adaption, Ganin2016} to removing unwanted biases in representation learning \cite{harrison2016censoring}. Similar to ours, a discriminator is trained to learn domain invariant feature representations for variety of tasks in \cite{GAN_domain_adaption}.

%% file: section/main.tex
\begin{figure*}
\begin{center}
\includegraphics[width=\linewidth]{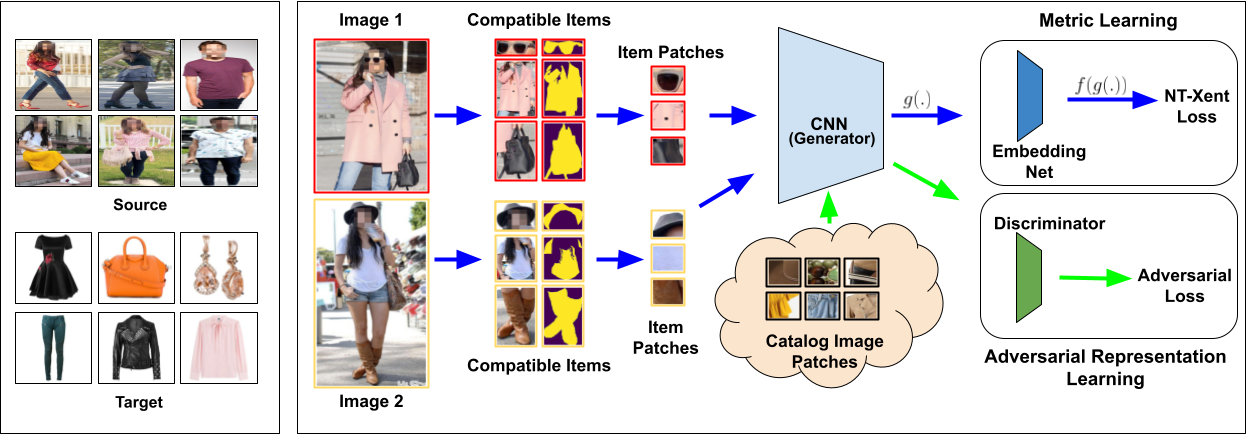}
\end{center}
\caption{Overview of our proposed self-supervised approach. It consists of a generator, discriminator and an embedding networks. Given a batch of in-the-wild from the source domain (left-top), we train the generator and embedding networks to learn representations on a pair of positive and negative patches from the same and different persons, respectively. Additionally, a discriminator is trained along with generator in a adversarial framework that adopts the model to the target domain (left-bottom).}
\label{fig:main}
\end{figure*}

People typically wear compatible outfits in their day-to-day lives in which different clothing pieces and accessories complement each other well. Such photos of people are easy to collect and abundant. Hence, in this work we aim to learn discriminative representations for fashion compatibility from in-the-wild images collected from the Internet as opposed to the labeled outfits in existing approaches \cite{eccv2018learning, cvpr2020fashion}. 

In this section, we first formulate our problem (section: \ref{section: sec_setting}) and then provide an overview of our network architecture (section: \ref{section: overview}). We then introduce our proposed self-supervised representation learning technique (section~\ref{section: self_supervision}) for compatibility prediction from unlabeled in-the-wild images. Finally, we describe our domain adaption technique (section~\ref{section: domain_adaption}) that is crucial to address the domain difference between in-the-wild (source) and catalog (target) images. 

\subsection{Setting}
\label{section: sec_setting}
We are given unlabeled images from source and target domains to train our model. Source images denoted as $D_s=\{(x^s_i)\}_{i=1}^{M}$ are the day-to-day images (examples are shown in Figure~\ref{fig:main} (left)) of people wearing different outfits in various illumination and background conditions. Outfits in these images contain diverse range of clothing and accessories ({\em e.g.} bag, sunglasses, etc) reflecting different fashion trends, seasonality and demographic regions. In this work, we assume that rough bounding boxes or segmentation masks around clothing regions are either available, or obtained by state-of-the-art segmentation algorithms \cite{modanet, jia2020fashionpedia}. 

Similarly, target images donated as $D_t=\{(x^t_j)\}_{j=1}^{N}$ are from product catalog (examples shown in Figure~\ref{fig:main} (left)) from the domain of our interest {\em e.g.} e-commerce application. Unlike street images, these images are captured in a controlled setting with high resolution, illumination and focused on a single product item. Given these unlabeled images ($\mathcal{X} = \{D_s, D_t\}$), our objective is then to learn a function $F(\Phi): \mathcal{X} \rightarrow \mathcal{\bar{X}}$ that maps any fashion image, independent of its domain, into an embedding space where compatible items are closer to each other compared to non-compatible items. Here, $\Phi = (\theta, \phi, \omega)$ are our model parameters.

\subsection{Network architecture}
\label{section: overview}
Our network architecture is shown in Figure~\ref{fig:main}. It follows adversarial learning framework \cite{goodfellow2014generative, simGAN, GAN_domain_adaption} consisting of a generator and discriminator networks along with an embedding network. The generator network is essentially a convolutional neural network (\textsc{CNN}) (e.g. ResNet-50) backbone with parameter $\theta$ that maps the inputs into mid-level intermediate visual representation. It accepts inputs from both source and target domains. The discriminator network with parameter $\phi$ is a two layer \textsc{MLP} with a single output node. It takes the intermediate representation from the generator as input and then classifies whether the representation belong to source or target domain. Similarly, the embedding block is a two layer \textsc{MLP} with parameters ($\omega$) converts the generator's representation into visual embeddings that are discriminative for compatibility task by minimizing a self-supervised loss function. 

\subsection{Patch-based representation}
\label{section: sec_patch_representation}
Most of the previous works \cite{eccv2018learning, cvpr2020fashion, iccv2019learning} employed image-based representations for compatibility prediction. However, in this work, we focus on patch-based representations similar to \cite{kim2020self} albeit for different reasons. First, unlike in similarity learning, shape information is not very helpful for compatibility where items from different categories needs to be matched. Second, as our goal is to learn fashion representation from real-world images, it helps to deal with unknown transformations of clothing items due to warping, rotation and difficulties induced due to overlay of one item over another (e.g. wearing jacket on shirt occludes shirt) or occlusion of the body parts. Given the bounding box with width $w$ and height $h$, we randomly extract a square patch of size $min(30, r\times min(w, h))$ where $r \in [0.1, 0.25]$ is a cropping parameter. Few patch examples are shown in Figure~(\ref{fig:main}). As we show in our experiments, such a patch-based representation achieve similar performance compared to image based representation.

\subsection{Self-supervised learning for compatibility}
\label{section: self_supervision}
Inspired by the recent success of self-supervised learning for visual representation \cite{gidaris2018unsupervised, noroozi2016unsupervised, noroozi2017representation, simclr, goyal2019scaling, berthelot2019mixmatch}, we aim to leverage unlabeled images for fashion compatibility. These approaches either define a pretext task (such as solving a puzzle \cite{noroozi2016unsupervised}, estimating image rotations \cite{gidaris2018unsupervised}) or employ data augmentation \cite{simclr} to generate positive and negative pairs for contrastive learning. In this work, we follow the latter and generate positive pairs from patches from different regions of the same person. The negative pairs are constructed from rest of the patches within the batch that belong to different people. Given the embeddings $f'(x^s_i)$ for a image patch $x^s_i$, we define our compatibility loss as
\begin{equation} l_{c}(\omega) = \sum_{\substack{(i,j) \\ i \ne j}} -log\frac{exp(\mathcal{S}(f'(x^s_i), f'(x^s_j))/\tau)}{\sum_{k\ne i} exp(\mathcal{S}(f'(x^s_i), f'(x^s_k))/\tau)}  \label{eqn:nxent} \end{equation}
where $(i,j)$ denote each positive pair (patches from different regions of the same person) within a batch. $\mathcal{S}(u, v) = u^t v / \|u\| \|v\|$ denotes cosine similarity between two vectors and $\tau$ is the temperature parameter. $f'(x) = f(g(x))$ denotes the final layer output feature of the embedding network where $g$ is the generator representation.

\subsection{Adversarial learning}
\label{section: domain_adaption}
While model trained with Eqn~\ref{eqn:nxent}. learns discriminative features for compatibility, it may not perform well for catalog images due to distribution shift between the two domains. To handle this, we jointly train the model for compatibility learning and domain adaption with the help of a discriminator network. The discriminator network is trained to distinguish the class label of the generator representation. Meanwhile, generator together with the embedding network tries to jointly learn to ``fool'' the discriminator by minimizing the difference in feature distribution between source and target domain, thereby producing embeddings that are useful for the compatibility task. 

Given a batch of source patches $x^s_i$, the generator network aims to produce feature representation that the discriminator is unable to identify correctly, and at the same time are useful for compatibility task. Hence, we optimize the objective function for the generator and embedding networks together as follows:
\begin{equation} L_G(\theta, \omega) = l_{c} - \lambda_{1} \sum_i log(D_{\phi}(g(x^s_i))) - \lambda_{2} \sum_j log(1-D_{\phi}(g(x^s_j)))  \end{equation}
where $g(\cdot)$ is the generator embedding and $l_c$ is the compatibility loss defined in Eqn~\ref{eqn:nxent} on network embeddings $f'(\cdot)$. The first term is minimized when the discriminator output is close to $0$ for the source patches.

Similarly, discriminator is trained with a batch of source ($x^s_i$) and target patches ($x^t_i$) with their labels defined as $0$ and $1$, respectively. The discriminator network is trained with binary cross entropy loss defined as:
\begin{equation} L_D(\phi) = -\sum_i log(D_{\phi}(g(x^t_i))) - \sum_j log(1-D_{\phi}(g(x^s_j))) \end{equation} 

As in the \textsc{GAN} framework \cite{goodfellow2014generative}, we follow a minimax training procedure where in the parameters of discriminator network, and generator together with embedding networks are updated alternatively. When updating the weights of discriminator, generator and embedding network weights are kept fixed and vice-versa.

%% file: section/datasets.tex
For training, we use Fashionpedia \cite{jia2020fashionpedia} dataset as a source of street fashion images and IQON3000 \cite{iqon3000} for domain adaptation. We do not use any labels from these datasets. Following previous works, our results are reported on two popular fashion compatibility datasets, namely Polyvore \cite{eccv2018learning} and Polyvore-disjoint \cite{eccv2018learning} outfits.

\textbf{Fashionpedia} 
consists of user uploaded $48K$ street-fashion photos collected from free license websites such as Unsplash, Kaboompics {\em etc}. These photos contain people wearing variety of clothes and accessories captured in different background, weather and camera conditions. We use readily available segmentation mask annotations for extracting our patches during training, however one could obtain such masks with state-of-the-art segmentation algorithms such as mask-RCNN~\cite{mask_rcnn} trained on a fashion parsing datasets \cite{modanet}. We highlight that obtaining fashion compatibility labels is harder than clothing segmentation which is a well studied problem.



\textbf{IQON3000} is a catalog dataset of fashion images comprising of around $670K$ fashion items. We use a random subset of these images (without any labels) to train our domain adaption model.


\textbf{Polyvore} is a crowd-sourced dataset containing outfits or sets of fashion items that complement each other. 
It consists of manually labeled $68K$ outfits that are split into $53K$, $10K$ and $5K$ into training, validation and testing sets, respectively. 

\textbf{Polyvore-disjoint} is a subset of the Polyvore dataset created by removing outfits that have common items between training, validation and testing sets. The dataset is challenging compared to Polyvore dataset and consists of $32K$ outfits. During inference, we use only product images patches and do not use the metadata associated with the products.


%% file: section/experiments.tex
\subsection{Implementation Details}
\label{sec:implementation_details}
We follow the same training and testing procedure as employed in previous approaches \cite{eccv2018learning, kim2020self} for a fair comparison. We use imagenet pre-trained ResNet-50 \cite{resnet} as our backbone network. Both the embedding and discriminator networks consists of two-layer fully connected layers with $64$ hidden and output units. We use a batch size of $32$ for both source and target batches.
We set $\lambda_{1}$ and $\lambda_{2}$ to $0.05$ and temperature parameter to $0.2$. The network is trained with stochastic gradient descent with an initial learning rate of $5e$-$5$ which is decayed by a factor of $0.015$ every $500$ steps. The validation set is used for tuning hyper-parameters and used early stopping to prevent over-fitting. Our approach is implemented with Pytorch and trained on a cluster with Nvidia 2080-Ti GPUs.

\subsection{Evaluation Tasks}
We evaluate our approach for {\bf Fill-In-The-Blank} (\textsc{FITB}) and {\bf compatibility prediction} (\textsc{COMP}) tasks \cite{eccv2018learning}. \textsc{FITB} is a question and answering task where the objective is to select the right choice among four candidate items given an incomplete outfit query. On the other hand, (\textsc{COMP}) is a binary task where the model has to predict whether the given outfit is compatible or not. We report the performance in overall accuracy and area under the curve (AUC) for \textsc{FITB} and (\textsc{COMP}) tasks, respectively. We consider the largest cosine similarity between average outfit and candidate item embedding to calculate the accuracy. Similarly, average pairwise distance between outfit items is used to compute the AUC.

\input{tables/all_tables}
 
\subsection{Comparison with state-of-the-art methods}
\label{sec:comparsion_sota}

\textbf{Self-supervised approaches:} We report the performance of popular self-supervised baselines \cite{noroozi2016unsupervised, gidaris2018unsupervised,  hinton1993autoencoders, zhang2016colorful, zhirong2018unsupervised, zhuang2019local} with similar \textsc{CNN} backbone architecture on the Polyvore and Polyvore disjoint datasets in Table~\ref{table:all-tables}. These numbers are taken directly from \cite{kim2020self}. Our method outperforms each of these baselines on both (\textsc{COMP}) and (\textsc{FITB}) tasks on both these datasets.

    
    
We also make a direct comparison with recently proposed \textsc{STOC}~\cite{kim2020self} which is based on three self-supervised tasks that learn colour, local texture and global texture representations. The proposed method outperforms each of the three sub-tasks independently, and has similar performance when compared to STOC. It is important to note that most of the existing self-supervised approaches consider train and validation images from the same source (Polyvore dataset). These models benefit from the dataset biases in achieving high performance. 

Finally, our proposed approach outperform STOC by a large margin in a cross-dataset setting where training and test images are from different domains (datasets). More importantly, our approach that is unsupervised, trained on street fashion and cross dataset images is able to achieve high performance without any label or domain information.

\textbf{Supervised approaches:} For completeness, we also compare our approach with previously reported supervised approaches - Siamese Network \cite{eccv2018learning}, Type-Aware Network \cite{eccv2018learning}, SCE-Net \cite{iccv2019learning} and CSA-Net \cite{cvpr2020fashion}. These methods are trained and tested on different splits of the Polyvore datasets. Our proposed cross-domain model performs better than baseline siamese network with triplet loss, and slightly under performs compared to latest approaches that use sophisticated architectures with attention networks. Our approach greatly reduces the performance gap between supervised and self-supervised approaches by leveraging easily available street fashion images that have different distribution compared to the target domain. 
    




\begin{figure}
\centering
\includegraphics[scale=0.45]{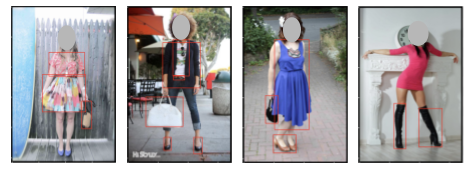}
\caption{Bounding boxes obtained by \textsc{YOLO} trained on Modanet dataset \cite{modanet}}
\label{fig:supp_inferred_bounding_boxes}
\end{figure}

\subsection{Ablation Studies}
\subsubsection{Training with predicted boxes}
As discussed in Section~\ref{section: sec_setting}, our approach requires bounding boxes/segmentation masks as weak labels to extract patches from different fashion items from each person. In Table~\ref{table:all-tables} we provided the results with ground-truth segmentation masks from the Fashionpedia \cite{jia2020fashionpedia}. With the advancements in object detection and availability of fashion datasets \cite{modanet}, it is however possible to automatically obtain these weak labels from pre-trained models. To demonstrate this, we conduct an additional experiment with detections obtained by the state-of-the-art YOLO-v3 detector \cite{yolov3} trained on the ModaNet dataset \cite{modanet}. We show a few example detections in Figure~\ref{fig:supp_inferred_bounding_boxes}. While training our model with these predicted bounding boxes, we construct positive pairs only for those images that have more than one detection. 


The model trained on these predicted bounding boxes achieves Compatibility AUC and FITB Accuracy of 0.83/54.9 and 0.81/53.2 on the Polyvore Outfits and Polyvore Disjoint datasets respectively. We notice only a minor drop in performance when comparing the model trained on the ground-truth which scores 0.84/55.5 and 0.82/54.6 making it suitable for large scale training.


\input{plots/tsne-adversarial}

\subsubsection{Role of Domain Adaption}
Table~\ref{adversarial-loss-table} shows the role of the proposed adversarial learning for fashion compatibility from cross-domain street images. It provides a consistent improvement of $1$-$3$\% on both tasks. We also show t-sne visualization of the source and target dataset embeddings without and with adversarial loss in Figure~\ref{adversarial-loss-tsne-normal} and Figure~\ref{adversarial-loss-tsne-adversarial}, respectively. It can be clearly seen that, our proposed adversarial training reduces the gap in feature distribution of two domains.

\subsubsection{Patch Representation}
To demonstrate the effectiveness of patch based representation, we train a fully supervised model \cite{eccv2018learning} with image and patch inputs and achieve compatibility and FITB performance of $0.84$/$54.6$ and $0.83$/$54.5$ respectively. We did not observe a significant drop in performance. 

\begin{figure}
\centering
\includegraphics[width=0.75\columnwidth]{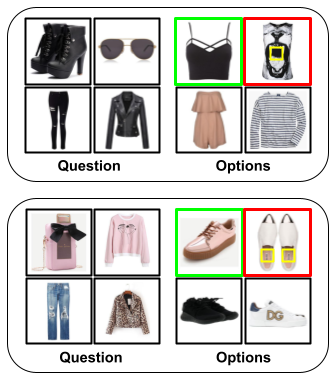}
\caption{Qualitative results showing success (green) and failure (red) cases for single and multiple patch based inference. Yellow box indicate regions that may not be the best representative of the item and when selected might produce incorrect predictions.}
\label{patches-ablation-qualitative}
\end{figure}

However, the location and size of patch extracted affects the performance. For example, patches marked with yellow border in Figure~\ref{patches-ablation-qualitative} are significantly different from the overall item, and such patches may degrade the performance. To avoid this, we extract multiple patches from an item, and consider the average score. 
Figure~\ref{fig:patches-ablation-compatibility} and Figure~\ref{fig:patches-ablation-compatibility}  shows the results for different number of patches during inference. We noticed a consistent improvement when more than $10$-$20$ patches are considered from each region beyond which there is no noticeable difference in performance.
Figure~\ref{patches-ablation-qualitative} shows a few qualitative examples where multiple patches improved the correct prediction rate.

Finally, as observed in \cite{kim2020self}, we notice a drop in the FITB and Compatibility prediction performance with increase in the patch size. Best performance is observed when the patch size is fixed to $min(30, r \times min(h, w))$ where $h$ and $w$ are the height and width of the item, respectively.



%% file: tables/all_tables.tex
\begin{table*}[h]
    \begin{minipage}{\dimexpr\linewidth-9.5cm\relax}
        \centering
        \subfloat[]{\label{table:all-tables}\input{tables/main_content}}
    \end{minipage}
    \begin{minipage}{0.5\linewidth}
        \centering
        \begin{minipage}{0.5\linewidth}
            \centering
            \subfloat[]{\label{adversarial-loss-table}\input{tables/adversarial_loss_content}} \par\medskip
            \subfloat[]{\label{fig:patches-ablation-compatibility}\input{plots/patches-ablation-compatibility}} \par\medskip
            \subfloat[]{\label{fig:patches-ablation-fitb}\input{plots/patches-ablation-fitb}}
        \end{minipage}
    \end{minipage}
\caption{(a) Comparison of our proposed approach on the Polyvore Outfits (left-top) and Polyvore Disjoint (left-bottom) datasets for the Compatibility Prediction and FITB tasks. Our approach achieves on-par performance when compared to a supervised siamese network and outperforms previous self-supervised approaches especially in challenging cross-dataset setting. (b) Effectiveness of the proposed adversarial training
for domain adaption. (c) \& (d) Effect of varying the patch size and number of patches during inference. Increasing the number of patches significantly improves the prediction performance on the Polyvore Outfits dataset. The plots also show performance comparison with different patch sizes. }
\label{table:main}
\end{table*}

%% file: tables/main_content.tex
\begin{tabular}{|c|c|c|c|}

\hline

\multicolumn{4}{|c|}{Polyvore Outfits Dataset} \\

\hline \hline

Category & Method & Compat. AUC & FITB Acc. \\

\hline \hline

\multirow{4}{2cm}{\begin{center}Supervised\end{center}} 
 & Siamese Network \cite{eccv2018learning} & 0.85 & 54.2 \\
 & Type-Aware Network \cite{eccv2018learning} & 0.86 & 55.3 \\
 & SCE-Net \cite{iccv2019learning} & 0.91 & 61.6 \\
 & CSA-Net \cite{cvpr2020fashion} & 0.91 & 63.73 \\

\hline \hline

\multirow{11}{2cm}{\begin{center}Self Supervised\end{center}} 
 & Jigsaw \cite{noroozi2016unsupervised} & 0.52 & 27.9 \\
 & Rotation \cite{gidaris2018unsupervised} & 0.53 & 29.4 \\
 & AutoEncoder \cite{hinton1993autoencoders} & 0.58 & 34.0 \\
 & Colorization \cite{zhang2016colorful} & 0.63 & 34.1 \\
 & ImageNet Weights & 0.66 & 39.1 \\
 & Instance Disc. \cite{zhirong2018unsupervised} & 0.74 & 45.9 \\
 & Local Agg. \cite{zhuang2019local} & 0.74 & 46.3 \\
 & RGB Histogram \cite{kim2020self} & 0.77 & 47.2 \\
 & Texture Descriptors \cite{kim2020self} & 0.77 & 50.3 \\
 & Patch Descriptors \cite{kim2020self} & 0.83 & 54.6 \\
 & STOC \cite{kim2020self} & 0.84  & 55.8 \\

\hline

\multirow{3}{2cm}{\centering + Cross Dataset} 
& \multirow{2}{*}{\centering STOC \cite{kim2020self}} & \multirow{2}{*}{\centering{0.81}} & \multirow{2}{*}{\centering{53.3}} \\
& \multirow{2}{*}{\centering Proposed Method} & \multirow{2}{*}{\centering{\bf 0.84}} & \multirow{2}{*}{\centering {\bf 55.5}} \\
& & & \\

\hline \hline

\multicolumn{4}{|c|}{Polyvore Disjoint Dataset} \\

\hline \hline

\multirow{3}{2cm}{\centering{Supervised}} 
 & Siamese Network \cite{eccv2018learning} & 0.85 & 54.4 \\
 & Type-Aware Network \cite{eccv2018learning} & 0.82 & 54.1 \\
 & CSA-Net \cite{cvpr2020fashion} & 0.87 & 59.26 \\

\hline \hline

\multirow{4}{2cm}{\centering{Self Supervised}} 
 & Instance Disc. \cite{zhirong2018unsupervised} & 0.69 & 43.2 \\
 & Local Agg. \cite{zhuang2019local} & 0.73 & 46.2 \\
 & RGB Histogram \cite{kim2020self} & 0.74 & 45.7 \\
 & STOC \cite{kim2020self} & 0.81  & 54.3 \\

\hline

\multirow{2}{2cm}{\centering{+ Cross Dataset}}
& \multirow{2}{*}{\centering{Proposed Method}} & \multirow{2}{*}{\centering{\bf 0.82}} & \multirow{2}{*}{\centering{\bf 54.6}} \\
& & & \\ 
 \hline

\end{tabular}

%% file: tables/adversarial_loss_content.tex
\centering
\begin{tabular}{|c|c|c|}
    \hline
    \multirow{2}{1.25cm}{Method} & \multicolumn{2}{|c|}{Polyvore Outfits} \\
    \cline{2-3}
        & Compat. AUC & FITB Acc. \\
    \hline
    $l_{c}$ & 0.82 & 54.4 \\
    $L_G + L_D$ & 0.85 & 55.4 \\
    \hline
    \multirow{2}{1.25cm}{Method} & \multicolumn{2}{|c|}{Polyvore Disjoint} \\
    \cline{2-3}
        & Compat. AUC & FITB Acc. \\
    \hline
    $l_{c}$ & 0.80 & 53.6 \\
    $L_G + L_D$ & 0.82 & 54.3 \\
    \hline
\end{tabular}

%% file: plots/patches-ablation-compatibility.tex
\begin{tikzpicture}[thick,scale=0.75, every node/.style={scale=0.75}]

\definecolor{color0}{rgb}{0.886274509803922,0.290196078431373,0.2}
\definecolor{color1}{rgb}{0.203921568627451,0.541176470588235,0.741176470588235}
\definecolor{color2}{rgb}{0.596078431372549,0.556862745098039,0.835294117647059}

\begin{axis}[
axis background/.style={fill=white!89.8039215686275!black},
axis line style={white},
legend cell align={left},
legend style={
  fill opacity=0.8,
  draw opacity=1,
  text opacity=1,
  at={(0.97,0.03)},
  anchor=south east,
  fill=white!89.8039215686275!black
},
tick align=outside,
tick pos=left,
x grid style={white},
xlabel={Number of Patches},
xmajorgrids,
xmin=-1.45, xmax=52.45,
xtick style={color=white!33.3333333333333!black},
y grid style={white},
ylabel={Compatibility AUC},
ymajorgrids,
ymin=0.78850855, ymax=0.843,
ytick style={color=white!33.3333333333333!black}
]
\addplot [very thick, color0, mark=*, mark size=3, mark options={solid}]
table {%
50 0.838363
40 0.8380670000000001
30 0.8377290000000001
20 0.8366790000000001
10 0.8346800000000001
5 0.829868
2 0.8161740000000001
1 0.795447
};
\addlegendentry{$r \in [0.10, 0.25], min = 30px$}
\addplot [very thick, color1, mark=*, mark size=3, mark options={solid}]
table {%
50 0.83497
40 0.834685
30 0.834559
20 0.834251
10 0.8333010000000001
5 0.830182
2 0.820422
1 0.8089149999999999
};
\addlegendentry{$r \in [0.25, 0.40], min = 50px$}
\addplot [very thick, color2, mark=*, mark size=3, mark options={solid}]
table {%
50 0.825092
40 0.824932
30 0.824653
20 0.824303
10 0.8217310000000001
5 0.816695
2 0.802931
1 0.808275
};
\addlegendentry{$r \in [0.40, 0.65], min = 80px$}
\end{axis}

\end{tikzpicture}

%% file: plots/patches-ablation-fitb.tex
\begin{tikzpicture}[thick,scale=0.75, every node/.style={scale=0.75}]

\definecolor{color0}{rgb}{0.886274509803922,0.290196078431373,0.2}
\definecolor{color1}{rgb}{0.203921568627451,0.541176470588235,0.741176470588235}
\definecolor{color2}{rgb}{0.596078431372549,0.556862745098039,0.835294117647059}

\begin{axis}[
axis background/.style={fill=white!89.8039215686275!black},
axis line style={white},
legend cell align={left},
legend style={
  fill opacity=0.8,
  draw opacity=1,
  text opacity=1,
  at={(0.97,0.03)},
  anchor=south east,
  fill=white!89.8039215686275!black
},
tick align=outside,
tick pos=left,
x grid style={white},
xlabel={Number of Patches},
xmajorgrids,
xmin=-1.45, xmax=52.45,
xtick style={color=white!33.3333333333333!black},
y grid style={white},
ylabel={FITB Accuracy},
ymajorgrids,
ymin=50.422, ymax=56.00,
ytick style={color=white!33.3333333333333!black}
]
\addplot [very thick, color0, mark=*, mark size=3, mark options={solid}]
table {%
50 55.47
40 55.3
30 55.21
20 55.3
10 55.03
5 54.39
2 52.66
1 50.7
};
\addlegendentry{$r \in [0.10, 0.25], min = 30px$}
\addplot [very thick, color1, mark=*, mark size=3, mark options={solid}]
table {%
50 54.8
40 54.66
30 54.72
20 54.5
10 54.59
5 54.15
2 52.93
1 52.0
};
\addlegendentry{$r \in [0.25, 0.40], min = 50px$}
\addplot [very thick, color2, mark=*, mark size=3, mark options={solid}]
table {%
50 54.14
40 54.14
30 54.23
20 54.1
10 54.02
5 54.05
2 52.65
1 52.49
};
\addlegendentry{$r \in [0.40, 0.65], min = 80px$}
\end{axis}

\end{tikzpicture}

%% file: plots/tsne-adversarial.tex

\begin{figure}[ht]
        \centering
        \begin{subfigure}{0.24\textwidth}    
        \centering
        \includegraphics[width=\columnwidth]{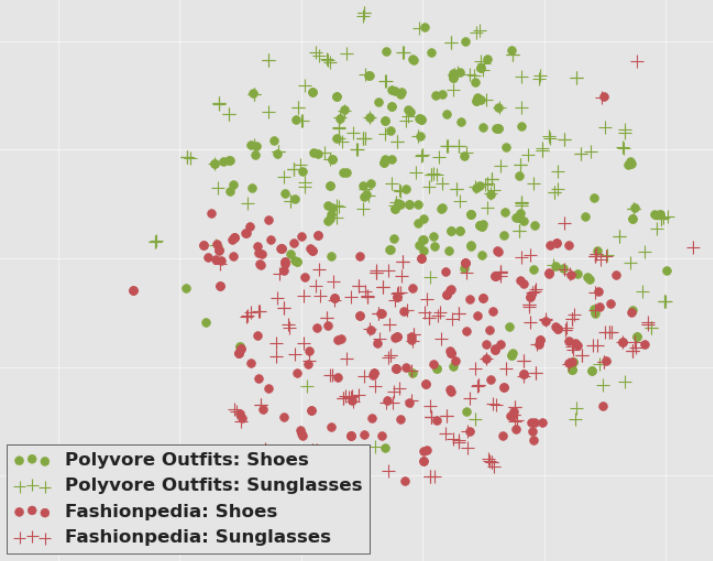}
        \caption{Without Adversarial Loss}
        \label{adversarial-loss-tsne-normal}
        \end{subfigure}
        \begin{subfigure}{0.24\textwidth}    
        \centering
        \includegraphics[width=\columnwidth]{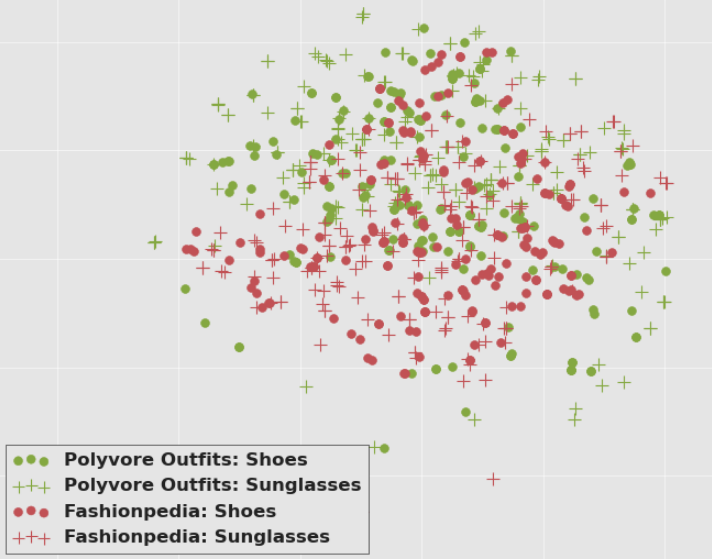}
        \caption{With Adversarial Loss}
        \label{adversarial-loss-tsne-adversarial}
        \end{subfigure}
        \caption{Effectiveness of the proposed adversarial training for domain adaption. We show the t-sne visualization of embeddings for two categories from the source and target datasets without (a) and with (b) domain adaption. Notice how the feature distribution for the two datasets is similar with adversarial training.}
\end{figure}

%% file: section/conclusion.tex
In this work, we present a self-supervised technique for learning visual representations for fashion compatibility prediction from unlabeled images. We generate positive and negative pairs from in-the-wild fashion items of the same and different person, respectively to train our model with contrastive learning. We further reduce the domain gap between source and target feature distributions by incorporating an adversarial loss. The experimental results suggest the representations learnt from unlabeled images are competitive to fully supervised methods. At the same time, our results are better than existing self-supervised approaches and outperform them by a significant margin in cross-domain evaluation settings.